# Transformer to GPT-5: Advancements, Limitations, and Future Directions in Large Language Model Design

1st Harsidak Singh Banwait
Chitkara University of Engineering and Technology,
Chitkara University,
Punjab, India
harsidak0343.becse25@chitkara.edu.in

2nd Chander Prabha
Chitkara University of Engineering and Technology,
Chitkara University,
Punjab, India
prabhanice@gmail.com

*Abstract*— **Large Language Models (LLMs) have revolutionized Artificial Intelligence (AI) by their extraordinary capabilities, like contextual understanding, few-shot/zero-shot learning, code generation, ability to reason and problem-solving. Recent studies have shown immense work on LLMs. Yet, there is limited work on the thorough taxonomy of LLMs. This study would showcase a detailed classification of LLMs, starting with the earlier models like BERT and leading to LLMs like GPT, Gemini, Claude and Grok. We would delve into topics like datasets used to train LLMs, computational power used during training, their hardware specifications, architecture and parameters. Further, we would also look at the limitations, computational challenges and open problems as well. This study would primarily focus on the timeline of LLMs, signifying their growth over the period. Findings reveal a clear pattern: models that optimize the ratio between architectural complexity and data quality exhibit stronger reasoning stability and better alignment, without proportionally increasing the computational load. This desired balance between scaling and sustainability is the defining characteristic of the current generation of AI.**

## I. INTRODUCTION

The Transformer model was released in 2017 [1]. Since then, LLMs have evolved from sequential and recurrent models to parallelizable and attention-driven learning frameworks that can describe long-range demands in text. The self-attention architecture of the Transformer model enabled models to pretrain on a massive scale, which is why they may have hundreds of billions of parameters. Adding additional parameters (such as GPT-3 with 175B) and training on vast volumes of unlabeled data have been associated with better performance and the advent of In-Context Learning (ICL) [2]. This enables you to complete tasks with no examples or very few examples during inference without needing to modify the parameters. A concise and practical taxonomy of LLM design elucidates the influence of architecture and size on Behaviour [3]. There are three primary methods for categorizing LLMs: Model size (the magnitude of its parameters), Type of architecture (decoder-only, encoder-only, or encoder–decoder), and type of input (text-only or multimodal with audio or images). When it comes to composing text, reasoning, or responding to queries, decoder-only models (such as GPT) are the most effective at generating open-ended content. Even though parameter scale still has a big effect on open-domain performance, things like how the architecture is set up, what kind of pretraining data is used, and fine-tuning methods like Parameter Efficient Fine Tuning (PEFT) methods like Low-Rank Adaptation (LoRA) [4] have a big effect on how useful, flexible, and efficient they are in the real world. Social media platforms are embracing the application of LLMs to censor content posted on their websites, provide suggestions, and offer automated customer services. The social media information, in turn, occupies a considerable portion of the training corpus of existing LLMs, casting doubt on the quality of information, bias reinforcement, and the reality that information created by AI is frequently reused to train subsequent models. The issue of misinformation, the echo chamber effect, and the truth of online conversations have been overstated due to the dissemination of content created by LLMs into social media.

## II. LITERATURE REVIEW

The progression of LLMs over the period states the rise of their training datasets and model parameters. As their training and datasets got more complex, their reasoning capabilities have significantly improved. However, including more datasets and increasing the complexity during the training of LLMs comes with certain challenges, and it makes it difficult to implement these models in real-life situations. As a result, considerable research has been conducted to enhance compression and optimization approaches, including quantization, pruning, and knowledge distillation. As LLMs transition from producing text to encompassing thinking, communication, planning, and multimodal cognition, we encounter complex challenges such as hallucination, bias, interpretability, and ethical alignment. These problems are at the heart of our discussions because they highlight the importance of AI systems being both reliable and transparent. This document aims to enhance existing debates by providing a repeatable chronology and a curated metadata corpus. A critical area for researchers in understanding how Large Language Models (LLMs) will impact society is to understand how these models function as social communication tools to produce & distribute (immediate) content, moderate content posted by users, & engage users. The knowledge of this impact should be established before any LLMs are used as a form of social communication, since many research projects have shown that LLM-generated content (training datasets) has significant effects on how an LLM behaves & how LLM-generated content affects the overall information ecosystem, as well as the challenges involved with monitoring this artificial content on a large scale.

## III. Methodology

This review comprises a reproducible metadata corpus and taxonomy of Large Language Models (LLMs) through a systematic search, manual curation, and dual validation [4]. The workflow consisted of the following steps: structured literature and model-card search across major repositories, explicit inclusion/exclusion screening, standardized metadata extraction into a CSV file, taxonomy construction via rule-based mapping, and validation through dual annotation. All data, extraction scripts, and plotting code are available at a public repository.

### *A. Search Strategy*

Searches were conducted on vendor/model-card pages (OpenAI, Meta, Hugging Face), arXiv, IEEE Xplore, ACL Anthology, NeurIPS/ICLR/ICML proceedings, and Google Scholar. The main subjects included "in-context learning," "RLHF," and "LoRA," while the main inquiries were "Transformer," "large language model," "LLM," "GPT," "BERT," and "LLaMA." The search was conducted between the years 2017 and 2024. Approximately 4,000 raw records were retrieved through searches, and from these, the most relevant papers were analyzed from March 3, 2025, to November 2, 2025.

### *B. Selection Criteria*

Inclusion requirements include primary sources (peer-reviewed articles, preprints, or official model-cards) that introduce or evaluate a model and report at least one quantitative description (e.g., parameter count) and historically significant releases, even if parameter counts are imprecise. Exclusion criteria include tutorial entries without primary metadata, duplicates, and toy-scale models (with fewer than 1 million parameters), unless they are of historical significance. Screening involved title/abstract screening, full-text assessment, and metadata extraction.

### *C. Taxonomy Construction*

The taxonomy design followed prior LLM classification frameworks [3]. The modality options are categorized along three axes: text, multimodal (combining vision and text), and audio. Architecture options include decoder-only, encoder-only, and encoder-decoder. Parameter scales considered for addition are small (less than 100M), medium (around 100M-(1B), big (approximately 1B-10B), XL (approximately 10B-100B), and HXL (greater than 100B).

## IV. Historical Timeline

The history of LLMs has witnessed numerous significant breakthroughs. The initial language models were intensive in statistical n-grams, which were limited in their contextual window size, and long-range dependencies were not taken into account. During the rise of LLMs algorithms like recurrent neural networks (RNNs) and long short-term memory (LSTM) contributed to their development significantly [4]. However, their stepwise computation lead to issues that become a barrier for the deployment of LLMs.

The release of the Transformer architecture in 2017 was a breakthrough in the sphere as it solved this problem. The GPT family was developed using an autoregressive design that facilitates the synthesis of high-quality text [5][6]. With these innovations, it is now possible to perform tasks that would normally be impossible, such as in-context learning (ICL), which enables models to detect patterns and complete tasks with minimal or no parameter adjustments. Complementary developments in alignment algorithms, particularly Reinforcement Learning from Human Feedback (RLHF), have enhanced the reliability of models that incorporate human-guided incentive schedules into their training pipelines. The rapid development of multimodal systems, which can work with text, vision, audio, and structured information, has broadened the strategic spectrum of LLMs, consolidating them into safer, more flexible, and functionally varied AI systems [7][8]. Table I presents a summary of various LLM architectures with their specifications. Table II presents a summary of the current multimodal LLM architectures.

## V. Challenges, Limitations, and Open Problems

The LLM knowledge representation, factual accuracy, and the larger training-deployment pipeline are still not up to par. These problems include both technical limitations and ethical and social risks.

### *A. Technical and Computational Challenges*

A major group of limitations comes from the huge size of these LLMs, and the problems arise while running them:
Computational and Energy Costs: It remains challenging to utilize LLMs on hardware that lacks sufficient resources or is low-end, as these models require substantial computational power. Both training and reasoning of LLMs require powerful GPUs, substantial RAM, and considerable engineering work. Their energy use is very high.

**Storage Requirements:** One of the main obstacles to during the development of LLMs is that they require a large amount of storage space to store data. It requires approximately 350 GB of storage space to store the FP16 weight matrices for models like GPT-3, which has 175 billion parameters [5]. As the parameters of models increase, their storage requirements become a significant challenge for deployment, replication, and large-scale testing.

**Scaling Law Limitations:** According to traditional scaling laws, increasing the size of such models and training them on higher-quality data should always improve their performance [9]. However, more recent studies suggest that these gains begin to diminish as models and datasets become increasingly large. This slowdown is known as sub-scaling behaviour.

Table I: Summary of LLM Architectures with their specifications

| Model/ Year | Architecture | parameters | Pre-trained data scale | Modality | Reported Compute | Weight Size Bytes | Training duration | Hardware Specifications |
|---|---|---|---|---|---|---|---|---|
| **BERT (Large)** 2018 | Encoder-Only | 340 M | Books Corpus + Wikipedia | Text | ~1 PF-day | 1.3 GB | Depends on model size | Nvidia A100, V100 |
| **RoBERTa(Large)** 2019 | Encoder-Only | 355 M | 160 GB Text | Text | – | 1.3 GB | ≈ 2 weeks | 6144 TPU v4 |
| **GPT-3** 2020 | Decoder-Only | 175 B | 300 B tokens | Text | $3.64 \times 10^{23}$ FLOPs | ~350 GB | ≈ 1 month | Nvidia A100 GPU |
| **T5-11b** 2020 | Encoder-Decoder | 11 B | 1 T tokens | Text | – | 44 GB | – | 1024 TPU v3 |
| **PaLM** 2022 | Decoder-Only | 540 B | 780 B tokens | Text + Code | $2.5 \times 10^{24}$ FLOPs | ~1.1 TB | 120 days | 6144 TPU v4 |
| **LaMDA** 2022 | Decoder-Only | 137 B | 768 B tokens | Dialogue | – | ~548 GB | 57.7 days | 1024 TPU v3 |
| **GLM-130B** 2022 | Decoder-Only | 130 B | 400 B tokens | Text | – | 520 GB | 60 days | Nvidia A100 |
| **Gopher** 2021 | Decoder-Only | 280 B | 300 B tokens | Text | – | 1.12 TB | 920 hrs | 4096 TPU v3 |
| **MT-NLG** 2022 | Decoder-Only | 530 B | 270 B tokens | Text | – | ~2.1 TB | – | 4480 A100 80 GB |
| **LLaMA 2** 2023 | Decoder-Only | 70 B | 2 T tokens | Text | – | 140 GB | 21 days | 2000 A100 80 GB |
| **Falcon** 2023 | Decoder-Only | 40 B | 1.3 T tokens | Text | – | 80 GB | – | Nvidia A100 |
| **Galactica** 2022 | Decoder-Only | 120 B | 106 B tokens | Scientific Text | – | 480 GB | – | 992 A100 80 GB |
| **BLOOM** 2022 | Decoder-Only | 176 B | 366 B tokens | Text (Multilingual) | – | 704 GB | 105 days | 384 A100 80 GB |
| **PanGu-α** 2021 | Decoder-Only | 207 B | 1.1 TB tokens | Text | – | ~828 GB | – | 2048 Ascend 910 |

Table II: Summary of the current multimodal LLM architecture

| Model/ Year | Architecture | parameters | Pre-trained data scale | Modality | Reported Compute | Weight Size Bytes | Hardware Specifications |
|---|---|---|---|---|---|---|---|
| **GPT-4** 2023 | Mixture-of-Experts (Decoder-Only) | ≈ 1 T (est.) | Undisclosed | Text + Image | Confi-dential | Undisclosed | Azure AI Supercluster |
| **Claude 3 Opus** 2024 | Transformer (Decoder-Only) | – (>100 B est.) | Proprietary Corpus | Text + Image | Confi-dential | – | Anthropic Cluster v2 |
| **Gemini 1.5** 2024 | Multimodal Transformer | 1.5 T | Web + Vision + Code Corpus | Text + Image + Code | ~1e24 FLOPs | ~3 TB | TPU v5e Pods |
| **LLaMA 3 70B** 2024 | Decoder-Only | 70 B | 15 T tokens | Text + Code | – | 140 GB | A100 80 GB |
| **DeepSeek-V3** 2024 | Mixture-of-Experts | 671 B (37 Active) | 14.8 T tokens | Text + Code + Reasoning | 2.788 M GPU hrs | ~1.34 TB | H800 GPU Cluster |
| **DeepSeek-R1** 2025 | Mixture-of-Experts (Reasoning) | 671 B (37 Active) | 14.8 T tokens | Text + Reasoning | – | ~1.34 TB | Confidential (H800 Cluster) |
| **GPT-5** 2025 | Unified system (fast + reasoning + router) | Undisclosed | Undisclosed | Text + Image (multimodal) | Un-disclosed | – | Undisclosed |
| **Gemini 2.5 Pro** 2025 | Multimodal Transformer / “Thinking” model | Undisclosed | Undisclosed | Text + Image + Audio/Video | Un-disclosed | – | Undisclosed |
| **Claude 4 (Opus)** 2025 | Advanced reasoning model | Undisclosed | Undisclosed | Text + Image (vision supported) | Un-disclosed | – | Undisclosed |
| **Grok 4** 2025 | Agentic multimodal reasoning model | Undisclosed | Undisclosed | Text + Reasoning + Tool Use | Un-disclosed | – | Undisclosed |

*B. Ethical, Societal, and Safety Issues*

Many individuals are concerned about the potential misuse of LLMs and the ethical issues that may arise due to their rapid proliferation.

**Bias and Discrimination:** LLMs might contain biases that could be present in their training data, worse, which could lead to unfair, negative, or biased results. The morals of today's models often reflect the values of Western or English-speaking groups. The training data, which is mostly extracted from the Internet, often contains false or harmful information [10][11].

**Privacy and Data Security:** Due to the popularity of LLMs and their active engagement, their risk for data privacy increases. During the collection of data in the pre-training stage, LLM providers usually collect their data from public internet sources such as literature, code repositories, and websites. LLM providers generally acquire publicly available data without the consent of the owner. While acquiring this data, it might include some sensitive information. Acquisition of this data through web scraping can lead to data leakage while potentially harming data protection standards [12].

**Misinformation and Social Media Amplification**: Due to the development of LLMs over 4 years, these models are successful in understanding context and exhibit reasoning that we can use these models for text generation and interpretation. Leveraging these capabilities of LLMs can be used for generating fake news and content. These models pose a threat to society, as individuals can use them to fake accounts and manipulate public opinion [13].

*C. Knowledge, Accuracy, and Generalisation Issues*

LLMs face limitations regarding the integrity and applicability of their learned knowledge:

**Hallucination:** In recent years LLMs have grown comprehensive reasoning capabilities, but we still ace with a major issue that is hallucinations. Sometimes LLMs can provide us with fluent and grammatically correct results but results can be contextually wrong [7]. These types of results can lead us to question reliability of LLMs as the output appears to be confident yet inaccurate.

**Generalization and Transferability:** It could be difficult for LLMs to provide desired results on specific topics related to a particular field. As most of these models are trained for general-purpose tasks and reasoning, they fail to present problem-solving abilities in depth for a specific area. These issues could arise due to limited access to data specific to that field. Somehow, alignment techniques like using small datasets and minimal retraining could improve their performance relative to the domain [14].

**Inaccuracy of Results:** Even after applying accurate training techniques and using quality data for training, LLMs can still produce inaccurate results [15].

*D. Training and Fine-Tuning Limitations*

While LLMs are generally trained for general tasks like writing, coding, etc. It remains challenging to train and tune LLMs for specific tasks in a way that aligns the results with human values and ethics.

**Training Complexity and fine-tuning**: One of the major problems with LLMs is that training these models requires a lot of time and money. It would require the acquisition of datasets to make these models work for specific tasks. It costs a lot of money and takes a lot of time to try out different training methods to see which ones work best.

**LLMs as Black-Box Models:** GPT and Gemini are two examples of LLM providers that work as black-box models. During the training, many techniques aim to match human values and require direct access to hidden states to understand what these models can interpret from the training data. These models are difficult to work on, making them less useful in environments where the source code is unavailable. [17].

**Reduced performance in quantization:** During the deployment of models, quantisation can be used to compress the size and memory required for LLMs by implementing low-precision weights. Performance of the model can significantly drop if compressed to an extent where the weights are compressed below 8 bits. Aggressive compression can be challenging to implement, as it compromises accuracy [18].

## VI. Open Problems and Future Research Gaps

Future research shall tackle the following fundamental issues that are still unsolved:

**Multimodal Integration:** While multimodal LLMs are still emerging, several challenges persist, including data explanation, the complexity of cross-modal accuracy and achieving sufficient reasoning capabilities across different domains (e.g., text, vision, audio) [19].

**Limits of Self-Correction:** The self-learning capacity of LLMs to reliably correct their reasoning and errors remain unresolved. Current studies indicate that self-correction in models and their resulting feedback is inconsistent and typically effective only on narrow tasks [20]. Broad, generalisable self-correction remains an open research frontier.

**Integration in social media and Adapting to Real-Time Conditions**: There are many challenges in developing LLMs that can adapt to rapidly changing social media. Future research might focus on developing methods for monitoring content on social media, detecting AI-generated images/content and establishing ethical guidelines for utilizing LLMs in the social domain [21].

## VII. CONCLUSION

The objective of the study is to delve into the timeline of LLMs beginning from the BERT model and to understand leading models like GPT, Gemini, and DeepSeek. This study also demonstrated how factors like architecture, parameters, and training data can significantly contribute to the scalability and deployment of these models. Furthermore, we also acknowledged certain limitations of LLMs, like hallucinations, etc., and how unethical usage of these models can be a threat to society. It could be difficult to deploy and train models in real time because of their increasing training data and computational requirements. The findings indicate that scaling LLMs leads to improved generalization and reasoning capabilities, including emergent behaviours such as in-context learning (ICL). Also, it indicates that we need to optimise according to hardware usage. Future work might focus on improving existing LLMs' ability to adapt in real-time to multimodal reasoning. Also making AI models that are consistent with human values, which would be beneficial for society.

## REFERENCES

[1] Y. Saleh, M. Abu Talib, Q. Nasir, and F. Dakalbab, "Evaluating large language models: a systematic review of efficiency, applications, and future directions," *Front. Comput. Sci.*, vol. 7, no. 1523699, 2025.

[2] H. S. Banwait, "LLM Evolution Metadata Repository," *Github*. [Online]. Available: https://github.com/Harsidak/LLM-Evolution-Metadata-Repository.git.

[3] S. Han, M. Wang, J. Zhang, D. Li, and J. Duan, "A review of large language models: Fundamental architectures, key technological evolutions, interdisciplinary technologies integration, optimization and compression techniques, applications, and challenges," *Electronics (Basel)*, vol. 13, no. 24, p. 5040, 2024.

[4] A. Sherstinsky, "Fundamentals of recurrent neural network (RNN) and long short-term memory (LSTM) network," *Physica D*, vol. 404, no. 132306, p. 132306, 2020.

[5] T. B. Brown *et al.*, "Language Models are Few-Shot Learners," *arXiv [cs.CL]*, 2020.

[6] J. Kaplan *et al.*, "Scaling laws for neural language models," *arXiv [cs.LG]*, 2020.

[7] Z. Ji *et al.*, "Survey of hallucination in natural Language Generation," *arXiv [cs.CL]*, 2022.

[8] P. Liang *et al.*, "Holistic Evaluation of Language Models," *arXiv [cs.CL]*, 2022.

[9] D. Liu *et al.*, "Large language models have intrinsic self-correction ability," *arXiv [cs.CL]*, 2024.

[10] J.-B. Alayrac *et al.*, "Flamingo: A visual language model for few-shot learning," *arXiv [cs.CV]*, 2022.

[11] A. Vaswani *et al.*, "Attention is all you need," *arXiv [cs.CL]*, 2025.

[12] W. X. Zhao *et al.*, "A survey of large language models," *arXiv [cs.CL]*, 2023.

[13] E. J. Hu *et al.*, "LoRA: Low-Rank Adaptation of large language models," *arXiv [cs.CL]*, 2021.

[14] E. M. Bender, T. Gebru, A. McMillan-Major, and S. Shmitchell, "On the dangers of stochastic parrots: Can language models be too big?," in *Proceedings of the 2021 ACM Conference on Fairness, Accountability, and Transparency*, 2021.

[15] N. Carlini *et al.*, "Extracting training data from large language models," *arXiv [cs.CR]*, 2020.

[16] A. Zou, Z. Wang, N. Carlini, M. Nasr, J. Z. Kolter, and M. Fredrikson, "Universal and transferable adversarial attacks on aligned language models," *arXiv [cs.CL]*, 2023.

[17] L. Weidinger *et al.*, "Taxonomy of Risks posed by Language Models," in *2022 ACM Conference on Fairness, Accountability, and Transparency*, 2022.

[18] N. Dziri *et al.*, "Faith and fate: Limits of transformers on compositionality," *arXiv [cs.CL]*, 2023.

[19] H. Li, L. Ding, M. Fang, and D. Tao, "Revisiting catastrophic forgetting in large language model tuning," *arXiv [cs.CL]*, 2024.

[20] L. Ouyang *et al.*, "Training language models to follow instructions with human feedback," *arXiv [cs.CL]*, 2022.

[21] T. Dettmers, A. Pagnoni, A. Holtzman, and L. Zettlemoyer, "QLoRA: Efficient Finetuning of Quantized LLMs," *arXiv [cs.LG]*, 2023.